# Code-Mixer Ya Nahi: Novel Approaches to Measuring Multilingual LLMs' Code-Mixing Capabilities

**Ayushman Gupta**[*1]    **Akhil Bhogal**[*2]    **Kripabandhu Ghosh**[2]

[1]South City International School, India   [2]IISER Kolkata, India

ayushman2346@gmail.com  akhilbhogal25@gmail.com  kripaghosh@iiserkol.ac.in

## Abstract

Multilingual Large Language Models (LLMs) have demonstrated exceptional performance in Machine Translation (MT) tasks. However, their MT abilities in the context of code-switching (the practice of mixing two or more languages in an utterance) remain under-explored. In this paper, we introduce Rule-Based Prompting, a novel prompting technique to generate code-mixed sentences. We measure and compare the code-mixed MT abilities of 3 popular multilingual LLMs: GPT-3.5-turbo, GPT-4, and Gemini Pro across five language pairs: English-{Hindi, Bengali, Gujarati, French, Spanish} using $k$-shot prompting ($k \in \{0, 1, 10, 20\}$) and Rule-Based Prompting. Our findings suggest that though $k$-shot prompting often leads to the best results, Rule-Based prompting shows promise in generating unique code-mixed sentences that vary in their style of code-mixing. We also use $k$-shot prompting to gauge the code-mixed to English translation abilities of multilingual LLMs. For this purpose, we create a gold-standard code-mixed dataset spanning five language pairs: English-{Hindi, Bengali, Gujarati, French, Spanish}. As a real-world application of our work, we create a code-mixed chatbot.

## 1  Introduction

Multilingual Large Language Models (LLMs) have shown remarkable performance in Machine Translation (MT) tasks (Brown et al., 2020). However, in bilingual and multilingual communities worldwide, speakers often alternate between two or more languages in a single utterance. This phenomenon is known as code-switching or code-mixing (Poplack, 2001). Several works have explored code-mixed text generation (Gupta et al., 2020a; Gautam et al., 2021; Mondal et al., 2022;

[*]Equal contribution.

Hsu et al., 2023), but code-mixed MT using Multilingual Large Language Models remains under-studied.

Yong et al. 2023 explore the abilities of multilingual LLMs in generating natural code-mixed conversations as well as code-mixed text on five general topics in five South-East Asian languages. Zhang et al. 2023 present an analysis of the code-switching abilities of multilingual LLMs across 4 tasks, including MT. In their study of code-switched MT, they perform 0-shot prompting on several LLMs to translate English sentences to Hinglish, and vice versa. Our work builds upon this study by assessing the code-switching abilities of newly-released, bigger multilingual LLMs such as GPT-4 and Gemini Pro using $k$-shot prompting across 5 diverse language pairs.

Besides $k$-shot prompting, we also explore rule-based generation of code-mixed sentences. Srivastava and Singh 2021 discuss the methods of Word-Aligned Code-Mixing (WAC) and Phrase-Aligned Code-Mixing (PAC) for rule-based generation of code-mixed sentences. We create four 'rule-based prompts' for code-mixed text generation using multilingual LLMs. Rule-Based Prompting marks a novel way to utilize LLMs for generating code-mixed sentences. As the name suggests, Rule-Based prompts enable generation of code-mixed sentences by manipulating the English reference sentence according to certain hard-coded Rules. By varying these Rules, we can obtain code-mixed sentences of different styles such as insertional and alternating (Poplack and Walker, 2003) without any in-context learning.

We report our results on our own dataset, which contains 600 gold-standard code-mixed sentences spanning five language pairs: English-{Hindi, Bengali, Gujarati, French, Spanish}.
Overall, our contributions are as follows:
(i) We study English to code-mixed and code-mixed to English translation abilities of multi-

| Language pair | English sentence | Gold-standard code-mixed sentence |
|---|---|---|
| English-Hindi | They put it back into the market. | Unhone ise market mein vapas daal diya. |
| English-Bengali | I have never been so wrong in all my life. | Ami amar life e kono din eto wrong hoi ni. |
| English-Gujarati | Regular refers to usual or normal things. | Regular thi tatpary chhe usual athva normal vastu. |
| English-French | It will I hope be examined in a positive light | Il sera, je l'espère, examiné sous un positive light. |
| English-Spanish | I have said why I support the Constitution. | He dicho por qué yo support la Constitución. |

Table 1: Examples of 5 English sentences and their corresponding gold standard code-mixed sentences for the English-{Hindi, Bengali, Gujarati, French, Spanish} language pairs.

lingual LLMs using $k$-shot prompting and fine-tuning (see Section 3).

(ii) We introduce a novel prompting method which we term as Rule-Based Prompting, to generate code-mixed sentences of varying styles (see Section 4).

(iii) We create and release a dataset[1] containing 600 gold-standard code-mixed sentences spanning five language pairs, i.e. English-{Hindi, Bengali, Gujarati, French, Spanish} (see Section 2.1).

(iv) To demonstrate a real-world application of our study, we create a code-mixed chatbot which can answer questions about our paper in the five language pairs we have explored (see Section 5).

## 2 Experimental Setup

### 2.1 Datasets

For each language pair, we use 100 randomly chosen English sentences from the datasets provided in Gupta et al. 2020b for our $k$-shot experiments. From the same dataset, we randomly choose 20 sentences as examples for few-shot prompting. We create gold-standard code-mixed sentences corresponding to these 120 sentences across five-language pairs, i.e. English-{Hindi, Bengali, Gujarati, French, Spanish}, which we use as reference sentences to evaluate the generations. In total, we have 600 sentence-pairs (120 per language pair). Examples of sentences from our dataset are presented in Table 1. Details about the annotation process can be found in Appendix B.2.

We use 0-shot prompting to investigate model performance for the code-mixed to English translation task on a real-world code-mixed Twitter dataset (Dhar et al., 2018) consisting of 100 randomly selected English-Hindi code-mixed sentences.

[1] The dataset is available at `https://tinyurl.com/cmdataset`

Zhang et al. 2023 cite Srivastava and Singh 2022 as the source of the dataset they use for their experiments on code-mixed MT, where the test dataset for English-Hindi to English (CM2M) and English to English-Hindi (M2CM) both contain 1500 sentences each. The code-mixed dataset for the English-Hindi to English translation task appears to be a Twitter dataset. It is to be noted that Zhang et al. 2023 describe the test dataset as containing 960 sentence pairs while we find it to contain 1500 sentence pairs. We choose 1495 sentences for English to English-Hindi, and 1382 sentences for English-Hindi to English translation.

For fine-tuning, we use PHINC dataset (Srivastava and Singh, 2020) which is a twitter dataset containing 13738 sentence pairs. We split the shuffled dataset into train and valid with train dataset having 11677 sentence pairs., while 2061 sentence pairs for validation.

### 2.2 Models

We run our $k$-shot prompting experiments on GPT-3.5-turbo, GPT-4, and Gemini Pro.

## 3 $k$-shot Prompting

| $k$-shot$_\alpha$ example | $k$-shot$_\beta$ example |
|---|---|
| "faculty ki quality atyant mahatvapoorn hai" | "the quality of the faculty is critical. : faculty ki quality atyant mahatvapoorn hai" |

Table 2: Comparison between a $k$-shot$_\alpha$ example and a $k$-shot$_\beta$ example.

We investigate the performance of LLMs on two tasks: English to code-mixed translation, and code-mixed to English translation, using four $k$-shot prompts, viz. 0-shot, 1-shot, 10-shot, and 20-shot.

For the English to code-mixed translation task, we experiment with two variants of one-shot and few-shot (10-shot and 20-shot) prompts: $\alpha$ and $\beta$. A $k$-shot$_\alpha$ prompt contains k examples of a code-mixed sentence. On the contrary, a $k$-shot$_\beta$ prompt contains as examples k pairs of an English sentence and its English-X code-mixed translation. Table 2 illustrates the difference between these two variants.

| Model | Experiment | English-Hindi | | | English-Bengali | | | English-Gujarati | | | English-French | | | English-Spanish | | |
|---|---|---|---|---|---|---|---|---|---|---|---|---|---|---|---|---|
| | | BLEU | R | M | BLEU | R | M | BLEU | R | M | BLEU | R | M | BLEU | R | M |
| GPT-3.5-turbo | 0-shot | 17.39 | 43.42 | 39.99 | 16.64 | 39.93 | 39.92 | 4.74 | 27.18 | 23.40 | 35.25 | 68.42 | 63.25 | 26.72 | 55.75 | 56.11 |
| | 1-shot$_\alpha$ | 26.33 | 54.54 | 54.72 | 17.5 | 43.96 | 41.53 | 10.78 | 32.51 | 34.84 | 31.87 | 66.45 | 61.17 | 25.65 | 54.09 | 54.09 |
| | 1-shot$_\beta$ | 25.64 | 52.64 | 53.49 | 16.95 | 43.94 | 40.70 | 11.52 | 35.15 | 38.16 | **35.68** | 68.12 | 63.42 | 23.24 | 54.79 | 52.91 |
| | 10-shot$_\alpha$ | 26.7 | 55.17 | 56.39 | 16.83 | 43.57 | 41.42 | 12.94 | 39.92 | 43.63 | 29.54 | 63.09 | 57.38 | 25.41 | 54.21 | 54.56 |
| | 10-shot$_\beta$ | 31.06 | 56.99 | 56.94 | 20.13 | 44.37 | 42.97 | 12.57 | 37.39 | 40.56 | 32.67 | 67.27 | 61.64 | 26.05 | 55.15 | 55.22 |
| | 20-shot$_\alpha$ | 28.15 | 54.41 | 55.87 | 18.57 | 44.10 | 43.08 | 13.90 | 36.16 | 41.01 | 33.44 | 65.86 | 61.2 | 26.06 | 54.36 | 54.74 |
| | 20-shot$_\beta$ | 31.27 | 56.92 | 56.22 | 18.42 | 44.58 | 42.51 | 13.65 | 38.75 | 42.02 | 33.58 | 67.05 | 62.05 | 26.57 | 55.51 | 56.41 |
| GPT-4 | 0-shot | 30.18 | 57.81 | 60.12 | **37.97** | 65.37 | 64.13 | 15.16 | 41.34 | 45.43 | 23.12 | 57.22 | 48.27 | 28.36 | 55.71 | 55.02 |
| | 1-shot$_\alpha$ | 32.95 | 61.22 | 63.95 | 36.67 | 64.23 | 62.51 | 16.89 | 43.74 | 46.93 | 15.69 | 50.83 | 42.8 | 21.25 | 54.8 | 48.1 |
| | 1-shot$_\beta$ | 35.2 | 63.25 | 64.95 | 30.64 | 58.69 | 57.61 | 18.16 | 45.10 | 46.61 | 16.1 | 45.49 | 36.57 | 27.04 | 56.24 | 55.16 |
| | 10-shot$_\alpha$ | 32.51 | 60.34 | 63.82 | 26.43 | 57.31 | 55.83 | 19.11 | 49.58 | 53.64 | 18.74 | 53.49 | 47.57 | 25.43 | 55.76 | 55.71 |
| | 10-shot$_\beta$ | 34.08 | 59.67 | 62.98 | 32.33 | 61.41 | 59.39 | **23.22** | 50.65 | 54.56 | 22.11 | 52.04 | 47.51 | 26.31 | 56.85 | 56.96 |
| | 20-shot$_\alpha$ | 34.01 | 61.23 | 64.87 | 26.58 | 57.82 | 55.96 | 20.01 | 48.27 | 52.99 | 18.41 | 53.91 | 48.21 | 25.58 | 58.8 | 58.77 |
| | 20-shot$_\beta$ | **36.65** | 61.34 | 65.23 | 30.24 | 59.43 | 57.14 | 21.94 | 50.44 | 54.41 | 24.52 | 56.15 | 51.85 | 25.28 | 55.58 | 56.13 |
| Gemini Pro | 0-shot | 26.88 | 57.25 | 54.71 | 20.54 | 44.58 | 44.03 | 7.35 | 29.34 | 28.87 | 19.78 | 52.34 | 45.83 | 22.93 | 52.62 | 52.09 |
| | 1-shot$_\alpha$ | 25.53 | 56.48 | 50.79 | 14.65 | 39.19 | 37.18 | 10.98 | 28.79 | 29.53 | 30.77 | 61.98 | 55.11 | 26.64 | 56.22 | 56.65 |
| | 1-shot$_\beta$ | 20.28 | 51.1 | 45.78 | 18.95 | 42.62 | 38.21 | 12.30 | 31.45 | 31.21 | 29.03 | 62.74 | 55.31 | 27.79 | 57.03 | 57.36 |
| | 10-shot$_\alpha$ | 25.11 | 55.91 | 50.49 | 19.83 | 46.78 | 41.32 | 9.48 | 29.02 | 30.73 | 32.74 | 62.66 | 57.16 | 31.56 | 58.7 | 58.01 |
| | 10-shot$_\beta$ | 25.17 | 58.35 | 50.37 | 17.69 | 45.26 | 39.5 | 10.76 | 30.83 | 30.67 | 34.72 | 64.71 | 59.91 | **36.97** | 63.57 | 62.46 |
| | 20-shot$_\alpha$ | 27.28 | 59.44 | 52.76 | 20.79 | 47.99 | 44.37 | 9.52 | 28.07 | 30.01 | 29.28 | 60.9 | 55.01 | 31.02 | 58.6 | 58.02 |
| | 20-shot$_\beta$ | 26.15 | 58.6 | 52.43 | 18.49 | 48.13 | 41.87 | 10.67 | 28.07 | 29.38 | 31.81 | 63.49 | 58.07 | 35.9 | 61.72 | 61.04 |

Table 3: English to Code-Mixed results. R: ROUGE-L (F1) score, M: METEOR Score.

#### 3.0.1 Testing on Zhang et al. 2023 Dataset

We run $k$-shot experiments for English-Hindi to English as well as for English to English-Hindi on this dataset. Due to budget constraints and the large size of the dataset, we use only Gemini Pro.

### 3.1 Fine-tuning

We fine-tune Flan-T5-Base (Chung et al., 2022) for the English to code-mixed translation task, on a Twitter code-mixed dataset (Srivastava and Singh, 2020).

### 3.2 Additional Explorations

We carry out prompting experiments with other open-source models, namely FLAN-T5-XXL (Chung et al., 2022), Mistral-7B (Jiang et al., 2023), and BLOOMZ-7.1B (Muennighoff et al., 2022), in both translation directions. However, due to consistently poor results across several attempts, we discontinue their use before evaluating their performance on our test data.

### 3.3 Results and Analysis

Before calculating metric scores, we clean noisy outputs containing tags or explanations, such as "Code-Mixed:" or "Transliteration to Roman:", using a combination of manual and automatic methods. In order to study the abilities of multilingual LLMs in generating code-switched text in the Roman script, we do not transliterate the non-Roman words generated by the models.

We use Corpus BLEU (using the NLTK [2] library) without the smoothing function, rouge-score 0.1.2[3]. We use the NLTK library to calculate METEOR score. We use a stemmer for the Code-Mixed to English experiments

### 3.3.1 English to Code-Mixed

Table 3 shows the BLEU, ROUGE-L (F1), and METEOR scores achieved by GPT-3.5-turbo, GPT-4, and Gemini Pro for the English to code-mixed translation task.

For the English-Hindi, English-Bengali and English-Gujarati language pairs, GPT-4 outperforms both GPT-3.5-turbo and Gemini Pro.

For English to English-Hindi, across all models and experiments, the maximum BLEU score we achieve is 36.65 for 20-shot$_\beta$ with GPT-4. For English-Bengali, we achieve a maximum BLEU score of 37.97 for 0-Shot with GPT-4.

For English-Hindi, GPT-4 achieves the highest scores for all $k$-shot experiments (k=0, 1, 10 and 20), followed by GPT-3.5-turbo, with Gemini-Pro being the worst performer.

---
[2] https://www.nltk.org/
[3] https://pypi.org/project/rouge-score/

As for English-Gujarati, we achieve a maximum BLEU score of 23.22 for 10-Shot$_\beta$ with GPT-4. GPT-4 outperforms both GPT-3.5-Turbo and Gemini-Pro by a significant margin across all experiments.

For English to English-Spanish, Gemini Pro achieves the best results across all experiments except 0-shot.

Gemini Pro achieves a maximum BLEU score of 36.97 for 10-shot$_\beta$.

For English-French, GPT-3.5-turbo performs the best, achieving a BLEU score of 35.68 across all experiments. GPT-4 achieves the lowest scores.

We do not observe a general pattern in the context of BLEU scores across $k$-shot experiments (B.3). However, we observe that $k$-shot$_\beta$ prompts frequently lead to better output than $k$-shot$_\alpha$ prompts for all language pairs except English-Bengali. This result leads us to infer that the inclusion of English and their corresponding code-mixed sentences in the prompt rather than only English-X code-mixed sentences, allows models to better understand the linguistic elements pertaining to code-mixing.

We note that for the English-Hindi, English-Bengali and English-Gujarati pairs, GPT-3.5-turbo frequently includes Devanagari, Bengali and Gujarati script tokens in its generations, despite specifying that the output must entirely be in the Roman script. This effect is most pronounced in the case of 0-shot prompting, and its frequency decreases as the number of examples increases. This observation leads us to conclude that GPT-3.5-turbo is incapable of consistently generating English-X code-mixed sentences in the Roman script, where X is non-Latin script language. We observe a marked decline in the performance of LLMs with respect to the English-Gujarati language pair. In the sentences generated by GPT-3.5-turbo, Gujarati script tokens appear more frequently than Devanagari script tokens in its English-Hindi generations or Bengali script tokens in its English-Bengali generations. Moreover, we find 13.90 to be the highest BLEU score for this task with respect to GPT-3.5-turbo. Therefore, we find that GPT-3.5-turbo is not an ideal choice for the English to English-Gujarati translation task.

We find that Gemini Pro is unable to translate many sentences, and often returns an empty output. Moreover, its translations are highly flawed. For instance, for the English sentence "Jawaharlal Nehru had no answer to this.", the 0-Shot translation Gemini Pro produces contains 684 words : "*Jawaharlal Nehru na aapne aapne aapne....*", where the word 'aapne' is repeated 681 times in default inference settings except for temperature set to 0. This suggests that Gemini Pro is not suitable for the English to English-Gujarati translation task. We find that GPT-4 performs the best and may be considered for this task. We attribute this dip in model performance to Gujarati being a low-resource language, and hypothesize that models may not have been trained on sufficient Gujarati data.

Results for fine-tuned Flan-T5-base are shown in Table 4. We achieve a BLEU score of 13.97 on the Zhang et al. 2023 (1495 sentences). The sentences generated are not up to the standard. This, we believe, is due to the size of dataset not being sufficiently large, further adding to the problem of the inherent spelling inconsistencies within the Romanized Hindi Twitter dataset.

| Dataset | BLEU | ROUGE-L (F1) | METEOR |
|---|---|---|---|
| Zhang et al. 2023 (1495 sentences) | 13.97 | 36.21 | 35.12 |
| Default Dataset (100 sentences) | 11.32 | 32.45 | 38.42 |

Table 4: Results for Code-Mixed English-Hindi to English for Fine-Tuned Flan-T5-Base

| Experiment | BLEU | ROUGE-L (F1) | METEOR |
|---|---|---|---|
| 0-shot | 15.00 | 41.74 | 41.58 |
| 1-shot$_\alpha$ | 14.27 | 41.33 | 39.19 |
| 1-shot$_\beta$ | 13.61 | 40.91 | 36.89 |
| 10-shot$_\alpha$ | 14.40 | 41.14 | 37.25 |
| 10-shot$_\beta$ | 14.82 | 41.49 | 37.77 |
| 20-shot$_\alpha$ | 15.60 | 42.80 | 39.89 |
| 20-shot$_\beta$ | **16.61** | 44.05 | 41.70 |

Table 5: English to code-mixed English-Hindi generation results for Gemini Pro on the dataset used in Zhang et al. 2023 (1495 sentences).

Table 5 shows the BLEU, ROUGE-L (F1), and METEOR scores that we achieve using Gemini Pro for the English to English-Hindi code-mixed translation task on the dataset used by Zhang et al. 2023. We achieve the highest BLEU score (16.61) using 20-shot$_\beta$ prompting. This score is significantly lower than the maximum BLEU score of 27.28 achieved previously on our test dataset for this task with Gemini Pro.

However, this difference in results may be attributed to the inefficacy of standard MT evaluation metrics such as the BLEU score in accurately evaluating code-mixed translations, as there may be several ways to translate an English sentence into an English-X code-mixed sentence. In addition, we observe that the reference sentences provided in the dataset are not consistently correct code-mixed translations. For instance, we consider the English sentence: "at the top of speech video". The gold-standard code-mixed reference sentence corresponding to this English sentence, as provided in the dataset is: "*stariya aur top of speech ka video*", which we believe to be an incorrect translation. On the other hand, translation generated by Gemini Pro (using a 0-shot prompt) is: "*speech video ke upar*", which we consider to be a correct English-Hindi translation of the original sentence.

| Model | Experiment | English-Hindi | | | English-Bengali | | | English-Gujarati | | | English-French | | | English-Spanish | | |
|---|---|---|---|---|---|---|---|---|---|---|---|---|---|---|---|---|
| | | BLEU | R | M | BLEU | R | M | BLEU | R | M | BLEU | R | M | BLEU | R | M |
| GPT-3.5-turbo | 0-shot | 49.91 | 76.4 | 80.5 | 43.96 | 66.33 | 71.29 | 27.68 | 57.37 | 61.12 | 43.23 | 79.48 | 76.7 | 73.87 | 89.18 | 90.71 |
| | 1-shot | 52.02 | 77.25 | 81.75 | 46.91 | 68.16 | 72.97 | 31.71 | 59.72 | 64.73 | 43.27 | 80.3 | 77.9 | 72.02 | 87.94 | 89.78 |
| | 10-shot | 46.14 | 73.63 | 79.9 | 45.75 | 68.24 | 74.13 | 31.65 | 62.24 | 66.28 | 42.46 | 80.09 | 77.71 | 73.41 | 87.95 | 89.98 |
| | 20-shot | 45.38 | 72.8 | 79.46 | **54.93** | 73.68 | 77.55 | 33.90 | 64.01 | 68.31 | 43.93 | 80.79 | 78.29 | 73.78 | 88.3 | 90.49 |
| GPT-4 | 0-shot | 57.9 | 80.56 | 83.6 | 53.1 | 71.34 | 76.3 | 38.52 | 68.96 | 73.47 | 43.91 | 80.59 | 78.48 | 73.23 | 88.45 | 90.14 |
| | 1-shot | 56.01 | 79.54 | 82.81 | 54.27 | 73.22 | 77.57 | 42.27 | 70.80 | 73.31 | 43.92 | 80.17 | 78.35 | 74.08 | 88.73 | 90.23 |
| | 10-shot | 54.76 | 80.32 | 83.43 | 54.72 | 73.84 | 77.63 | 41.65 | 71.54 | 75.55 | 51.23 | 82.27 | 81.66 | **75.42** | 89.25 | 90.58 |
| | 20-shot | 54.72 | 79.95 | 83.57 | 54.67 | 73.71 | 77.02 | **42.62** | 72.76 | 77.05 | 50.68 | 83.15 | 82.31 | 75.41 | 89.83 | 90.99 |
| Gemini Pro | 0-shot | 60.24 | 81.78 | 82.52 | 45.13 | 66.81 | 71.01 | 29.77 | 56.56 | 60.70 | 53.13 | 80.56 | 79.07 | 72.49 | 88.52 | 86.44 |
| | 1-shot | **60.35** | 81.82 | 79.89 | 45.44 | 66.2 | 70.01 | 36.64 | 64.05 | 67.50 | 51.27 | 79.76 | 78.41 | 70.2 | 87.15 | 83.87 |
| | 10-shot | 56.79 | 79.9 | 79.16 | 49.49 | 68.89 | 71.85 | 38.32 | 65.40 | 68.31 | 54.94 | 82.3 | 81.17 | 74.4 | 89.54 | 88.08 |
| | 20-shot | 56.86 | 80.57 | 77.12 | 46.43 | 67.33 | 70.19 | 39.32 | 67.58 | 68.31 | **55.05** | 82.25 | 81.25 | 74.96 | 89.07 | 87.89 |

Table 6: Code-mixed to English ; Results for Code-Mixed to English translation for four language pairs. R is ROUGE-L (F1) score and M is the METEOR Score. For BLEU, Corpus BLEU has been calculated.

### 3.3.2 Code-Mixed to English

Table 6 shows the BLEU, ROUGE-L (F1), and METEOR scores achieved by GPT-3.5-turbo, GPT-4, and Gemini Pro for the code-mixed to English translation task.

Gemini Pro performs the best for English-Hindi and English-French across all $k$-shot (k=0, 1, 10, 20), followed by GPT-3.5-turbo, and then Gemini without any exceptions.

For English-Hindi, we achieve a maximum BLEU score of 60.35 for 1-shot with Gemini Pro. For English-Bengali, GPT-3.5-turbo achieves the highest BLEU score (54.93) for 20-shot prompting. For English-Gujarati, 20-Shot with GPT-4 achieves the highest BLEU score of 42.62. In the case of English-French, Gemini Pro achieves a maximum score of 55.05 for 20-shot prompting. For English-Spanish, we achieve the highest BLEU score (75.42) with GPT-4 using 10-shot prompting.

We observe that in the case of English-Spanish, there is not much variation in scores across models. However, for English-Bengali, GPT-4 demonstrates the best performance on average across experiments, despite GPT-3.5-turbo achieving the highest BLEU score. For English-Gujarati, GPT-4 performs the best across all $k$-Shot experiments, followed by Gemini Pro. Going through the outputs for English-Gujarati pair, we find that there are some wrong translations, which we can find in all the models. For English-Gujarati, GPT-3.5-turbo and GPT-4 suffer the most with using wrong pronouns. Other than that, we also find hallucination in some outputs as well, which is mostly found in GPT-3.5-turbo, followed by Gemini Pro. Thus, we do not find the LLMs completely reliable for English-Gujarati to English translation.

Table 7 shows the results of code-mixed English-Hindi to English translation, using 0-shot prompting, for 100 code-mixed sentences selected randomly from the Twitter dataset.

| Model | BLEU | ROUGE-L (F1) | METEOR |
|---|---|---|---|
| GPT-3.5-turbo | 37.95 | 65.55 | 69.06 |
| GPT-4 | 41.17 | 67.99 | 71.36 |
| Gemini Pro | **45.44** | 66.79 | 70.26 |

Table 7: Results for Code-Mixed English-Hindi to English Results 100 Twitter sentences. We perform only 0-shot prompting.

We achieve a maximum BLEU score of 45.44 with Gemini Pro. While this is significantly lower than the previous maximum of 60.35 for the goldstandard English-Hindi to English task, we observe that the models have no issues in understanding the Twitter data, and are often not confused by words written in short forms and spelling varia-tions.

Table 8 shows the BLEU, ROUGE-L (F1), and METEOR scores achieved by Gemini Pro for the code-mixed English-Hindi to English translation task on the dataset used by Zhang et al. 2023.

| Experiment | BLEU | ROUGE-L (F1) | METEOR |
|---|---|---|---|
| 0-shot | 29.84 | 58.65 | 55.77 |
| 1-shot | 29.45 | 58.40 | 55.99 |
| 10-shot | 30.13 | 58.90 | 56.96 |
| 20-shot | **30.20** | 58.70 | 56.89 |

Table 8: English-Hindi to English translation for (Zhang et al., 2023) dataset (1382 sentences) on Gemini Pro

We achieve a maximum BLEU score of 30.20 for 20-shot prompting.

It is to be noted that Gemini Pro removes from the output the mentions (@username) and URL links present in the input sentence. Since the dataset seems to be a Twitter dataset, many code-mixed sentences in the dataset, which we use as reference sentences for evaluation, contain such mentions and URL links. This may have led to the scores being significantly lower from the scores that we report for previous experiments.

## 4 Rule-Based Generation

We draw inspiration from previous work on synthetic rule-based generation of code-mixed sentences (Srivastava and Singh, 2021) to propose a prompt-based implementation of rule-based generation. We propose *four* 'rules' to generate code-mixed sentences, and integrate them into prompts, which we term as 'rule-based prompts'. To our knowledge, this marks the first attempt at leveraging 'rule-based' prompts for the generation of code-mixed sentences.

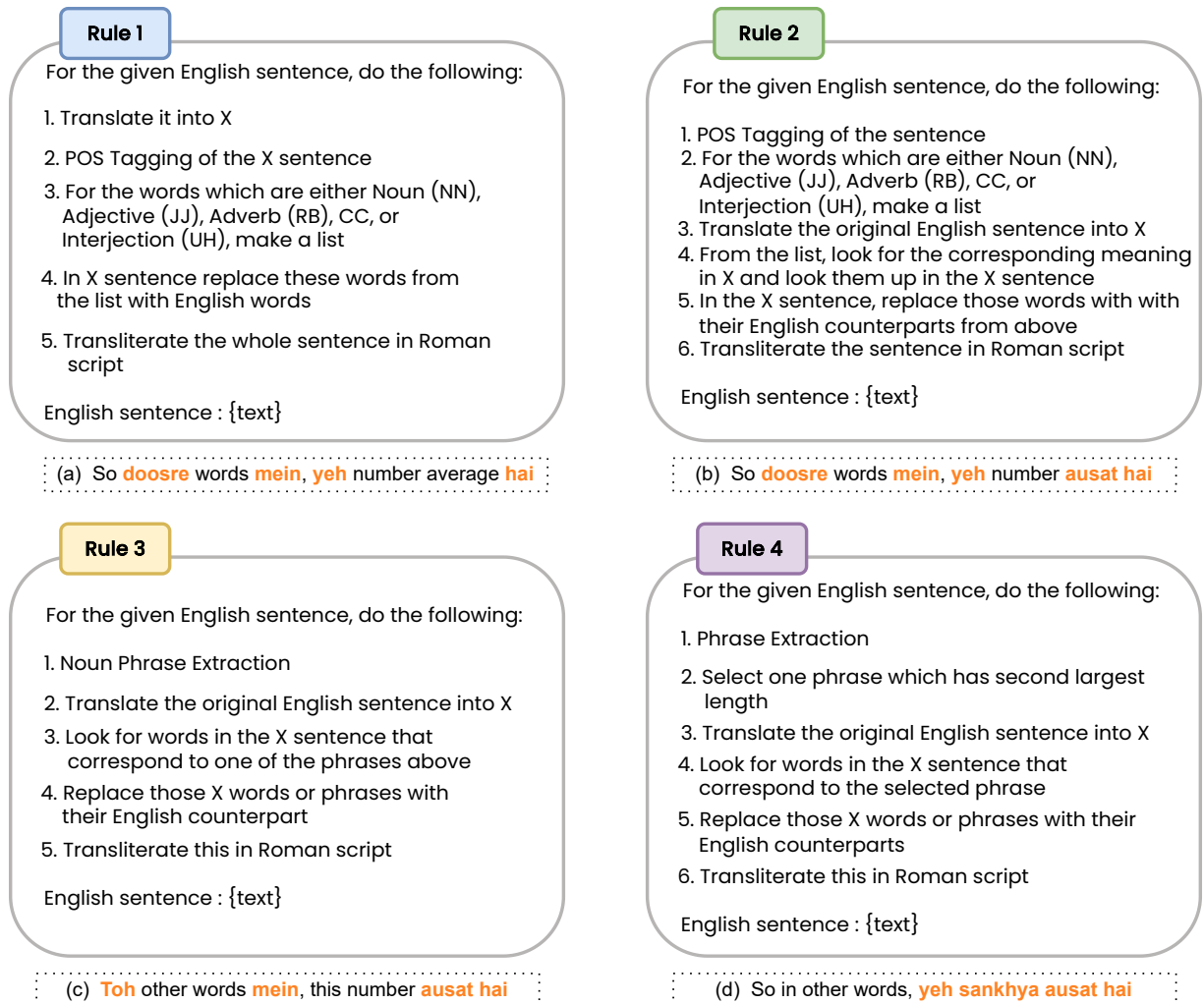

Figure 1: Rules for Rule-Based Generation with English-Hindi code-mixed examples

Through our rule-based prompting approach, we intend to generate code-mixed sentences from given English sentences, taking English to be the embedded language. This approach leverages the ability of the model to perform NLP tasks such as PoS tagging and to follow instructions in a sequential format to arrive at an output. While rule-based prompting may sometimes lead to the generation of noisy code-mixed sentences containing spelling and grammatical errors, we note that the generation of noisy code-mixed sentences is essential for studying the efficacy of evaluation metrics in accurately evaluating code-mixed generations across a spectrum of quality.

Figure 1 describes the four rules.

### 4.1 Rules 1 and 2

In Rule 1, the English sentence is first translated to the matrix language, which is PoS tagged. We choose words associated with specific PoS tags to replace with their English counterparts.

However, we find that PoS tagging is not consistently accurate for non-English tokens. To overcome this issue, we introduce Rule 2, which incorporates PoS tagging of the English sentence instead.

### 4.2 Rules 3 and 4

To generate code-mixed sentences that differ stylistically from the ones obtained using Rule 1 and Rule 2, we propose Rules 3 and 4. In Rules 1 and 2, specific PoS tagged words from the matrix language translation of the English sentence are replaced with their English counterparts to obtain a code-mixed sentence. To vary the style of code-mixing, we include English phrases in the code-mixed sentence instead of only singular words.

In Rule 3, noun phrases in the English sentence are identified. Subsequently, in the Hindi (translated) sentence, the corresponding parts are replaced with these English noun phrases.

Rule 4 divides the sentence into multiple phrases. We choose the phrase with the second-largest length to be in English. As per our observations, this leads to the generation of more natural and higher quality code-mixed sentences. This choice is subjective.

### 4.3 Results and Analysis

Rule 1 and Rule 2 perform the best, and Rule 2 frequently outperforms Rule 1. Rule 3 and Rule 4 consistently achieve lower BLEU scores than Rules 1 and 2. Our analysis suggests that rule-based prompting is a reliable approach to generate stylistically dissimilar code-mixed sentences from a given English sentence.

We observe that the results are not always ideal. With Rules 1 and 2, some generations are slightly unnatural and grammatically incorrect. For example, for the English sentence "*And the answer is, it depends.*", one of the generations is: "*And uttar hai, yeh depends karta hai*". The verb 'depends' is not aligned with the Hindi part of the compound English-Hindi verb ('*karta hai*': 'to do [something]'). This issue arises because we specify in the prompt that the words in the Hindi sentence must be replaced with their English translations from the English sentence.

Out of all the rules, Rule 3 leads to the generation of the most unnatural and grammatically incorrect sentences. Contrarily, Rule 4 often allows for the generation of grammatically correct, stylistically different code-mixed sentences. One example is "*Main from this entire thirty ek udhaar leta hoon*". However, some grammatically incorrect and unnatural sentences are also generated. One such example is "*The questions chaar prakaar ke hote hain*": in this sentence, the inclusion of the article 'the' makes the sentence semantically redundant, as this word has no direct translation in Hindi. Another example is "*Lekin aise avsar should be few aur door hone chahiye*". In this example, "hone chahiye" is superfluous as it conveys the same meaning as "should be". To explore diverse model capabilities for rule-based generation, we initially tested GPT-3.5-turbo and Gemini Pro alongside GPT-4 in the English to Hindi-

| Rule | English-Hindi | | | English-Bengali | | | English-Gujarati | | | English-French | | | English-Spanish | | |
|---|---|---|---|---|---|---|---|---|---|---|---|---|---|---|---|
| | BLEU | R | M | BLEU | R | M | BLEU | R | M | BLEU | R | M | BLEU | R | M |
| Rule 1 | **24.43** | 54.95 | 54.51 | 16.23 | 40.73 | 42.33 | **13.14** | 37.13 | 36.45 | 22.92 | 55.2 | 49.33 | 19.65 | 48.77 | 49.98 |
| Rule 2 | 23.09 | 53.0 | 53.19 | **18.01** | 44.04 | 45.26 | 8.75 | 32.06 | 29.58 | **24.31** | 54.86 | 48.5 | **24.16** | 51.11 | 50.77 |
| Rule 3 | 21.26 | 50.77 | 52.71 | 14.33 | 35.89 | 38.06 | 8.63 | 34.52 | 35.39 | 13.81 | 43.94 | 38.59 | 13.25 | 46.58 | 44.97 |
| Rule 4 | 12.75 | 42.43 | 43.78 | 11.85 | 30.1 | 33.95 | 7.48 | 28.26 | 27.36 | 18.96 | 46.0 | 41.27 | 15.03 | 42.11 | 42.64 |

Table 9: Results for Rule-Based Generations for the four language pairs ; R : ROUGE-L (F1) and M : METEOR.

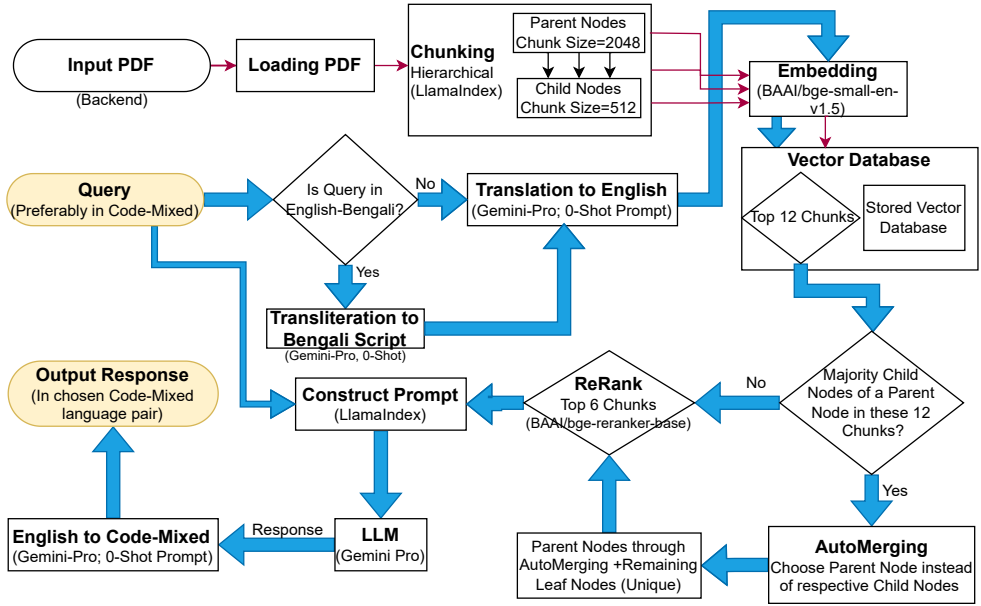

Figure 2: Flow of ChatBot

English translation task. While all three models underwent testing with a few sample sentences, GPT-3.5-turbo and Gemini Pro did not produce the desired level of quality or align well with the rule-based approach. Consequently, we opted to focus solely on GPT-4 for this specific task due to its demonstrated compatibility and superior performance in our initial evaluations.

## 5 Code-Mixed Chatbot

As an application of our work, we use Retrieval-Augmented Generation (RAG) to develop a Chatbot capable of answering code-mixed questions about our paper in any of the five code-mixed language pairs we have experimented with. Figure 2 explains the working of Chatbot. We use Gemini Pro as the LLM.

Appendix C.3 contains technical details about the chatbot's working as well as examples of queries and the chatbot's outputs.

## 6 Conclusion

In this paper, we first study the capabilities of GPT-3.5-Turbo, GPT-4 and Gemini-Pro in generating code-mixed sentences across five language pairs using $k$-shot prompting and fine-tuning. Using $k$-shot prompts, we also study how LLMs perform on the converse task, i.e. code-mixed to English translation. We do not observe a stark general trend in the results and infer that the results do not vary with respect to the prompting method in a consistent way. Therefore, the right combination of model and method is required to achieve the best results, as each model shows superior performance for at least one language pair. We observe that multilingual LLMs generally are capable of generating high quality code-mixed sentences, except when English is paired with a low-resource language such as Gujarati. Additionally, having defined and experimented with two kinds of in-context examples, viz. $\alpha$ and $\beta$ for the English to code-mixed task, we find that $k$-shot $\beta$ prompts consistently perform better than $\alpha$ prompt. We infer that presenting in-context examples as pairs of the English sentences and its code-mixed translation leads to better results.

Further, we introduce four rules for generating code-mixed sentences, and integrate them into prompts. We find that although $k$-shot prompting generally outperforms rule-based prompting, this approach shows promise in generating code-mixed sentences of unique styles. We also introduce a new code-mixed dataset containing 600 gold-standard code-mixed sentences spanning five language pairs, on which we report our results. As an application of our work, we develop a code-mixed chatbot which answers questions about our paper in any of the five language-pairs we have experimented with in our study.

# Limitations

The gold-standard sentences used for $k$-shot prompting were created by human annotators. Due to practical considerations, only 120 gold-standard code-mixed sentences were created for each language pair. Therefore, while considering the variety of prompting techniques and experiments we have tried, we test the models' performance on 100 sentences only for each language pair.

# Ethical Considerations

The human annotators are experts in their respective languages who have been paid commensurate to their efforts. The paid models have been used upon subscription.

## A Further Analysis of 'Code-Mixed to English' Results for English-Bengali and English-Gujarati

As have been discussed before, the performance of models for English-Gujarati to English is subpar, especially for GPT-3.5-Turbo and Gemini-Pro. Take the following as an example for English-Gujarati, for Gemini-Pro, where same code-mixed sentence yields English translations through 0-Shot and 1-Shot that mean the opposite of each other :
0-Shot: "Hold your body away from the phone."
1-Shot: "Keep your phone away from your body."

Another example is: "electrolytes that we want." (Original English)
0-Shot: "You should take Electrolyte."
1-Shot: "You need electrolytes."
10-Shot: "Electrolyte that you need."
20-Shot: "Electrolyte that you need."

This is worse for GPT-3.5-Turbo.
Correct English Sentence: "They and their spouses in shades reclining on couches."
0-Shot: "Teo and their spouses are reclining on the couches above the shades."
1-Shot: "Teo and their spouses reclined on the couches under the shades."
10-Shot: "Teo and their spouses recline on the couches in shades."
20-Shot: "His and her spouses recline on the couches in shades."

For English-Bengali, the sentence "finetuning er somporke bolo"(Tell me about finetuning) is wrongly translated by Gemini-Pro to "finetuning is always good"

For English-Gujarati, even GPT-4 has 'Teo' in some of its translations. It is to be noted that the temperature for these experiments is 0 and thus these different outputs do not result from that. Interpreting these results, it is likely that these models struggle especially when dealing with the Romanized Bengali and Gujarati, considering that we see words like 'Teo' in the translation. This can be because these languages are not high-resource languages like Hindi, and their Romanized dataset on which these models may have been trained, i.e. Romanized English-Bengali and English-Gujarati may not be sufficiently large. To test our hypothesis, we first transliterate these sentences into Bengali or Gujarati Scripts respectively and then translate into English with 0-Shot Prompting. We see some improvements in the translations, especially for Gemini-Pro.
For 'finetuning er somporke bolo', we get the Gemini-Pro output : 'Tell me about fine-tuning' which is correct.
"electrolytes that we want."(Correct English) : "Electrolytes that we need."(Gemini-Pro)

The problem of 'Teo' is also solved in GPT models. Quantitative Results are as follows:

| Model | English-Gujarati | | | English-Bengali | | |
|---|---|---|---|---|---|---|
| | BLEU | METEOR | ROUGE-L(F1) | BLEU | METEOR | ROUGE-L(F1) |
| Gemini-Pro | 43.88 | 73.27 | 71.79 | 53.87 | 76.08 | 72.72 |
| GPT-3.5-Turbo | 27.66 | 62.21 | 57.25 | 38.58 | 68.39 | 62.90 |
| GPT-4 | 37.97 | 72.71 | 68.61 | 43.93 | 71.43 | 66.26 |

Table 10: 'Code-Mixed to English' by first transliterating Code-Mixed to the Matrix Language's script and then translating that to English

We observe that Gemini-Pro is not capable of transliterating English-Gujarati in Roman script to Gujarati script, where it doesn't give any output for most sentences and thus, transliterations from GPT-3.5-Turbo have also been used for further translation step for Gemini-Pro. For English-Gujarati, though $k$-Shot prompting, the highest BLEU score achieved by Gemini-Pro is 39.32, while for GPT-3.5-Turbo and GPT-4, it is 33.90 and 42.65 respectively,

For English-Bengali, though $k$-Shot prompting, the highest BLEU score achieved by Gemini-Pro is 49.49, while for GPT-3.5-Turbo and GPT-4, highest BLEU scores are 54.93 and 54.72 respectively,

It can be seen from the results that Gemini-Pro's translations improve significantly, where for English-Gujarati, they beat the previous overall best BLEU score of 37.97 achieved by GPT-4.

But for GPT-3.5-Turbo and GPT-4, the scores worsen, even though some problems like 'Teo' word in English translation are fixed.

It is to be noted that, even though we get such improved scores for Gemini-Pro, this method is to be used cautiously as with transliteration, some new errors can be encountered. Take for example the following:
Gold-Standard English Sentence: "Bhagat Singh, however, was not supportive of severe violence."
Output of Gemini-Pro by this method: "Bhagat Singh, however, was not supportive of service violence."
0-Shot: "Bhagat Singh, who was, not supportive of violence."
1-Shot: "Bhagat Singh, who was not supportive of violence."
10-Shot: "Bhagat Singh, who was not supportive of violence."
20-Shot: "Bhagat Singh, who was not supportive of violence."

The word 'service' was never there in the previous translations and thus is a possible consequence of transliterating to Gujarati script.

This experiment was done to prove our hypothesis that these models (especially Gemini-Pro) struggle more due to the text being in roman script for translation to English and the outputs through this method while being better in one way, may suffer from another problem which was not even there before.

## B Supplementary Information

### B.1 $k$-shot Prompts

---

**0-shot**

Generate Code-mixed Bangla-english translation for : <Sentence>
The output should be a single sentence in roman script.

---

**1-shot$_\beta$**

Here is an example of translation of English sentence into code-mixed Bangla-English sentence:

English: "The trick is to start to build right from the back of your throat"
Code-Mixed: "Trick ta holo to start to build ekdom tomar throat er pechon theke"

Provide the code-mixed Bangla-English translation for the given English sentence: <sentence>
The output should be in Roman and shouldn't contain the original English sentence or any other tag.

---

**1-shot$_\alpha$**

Here is an examples of code-mixed Bangla-English sentence : "Trick ta holo to start to build ekdom tomar throat er pechon theke"

Provide the Code-mixed Bengali-English translation for the given English sentence.

The output should be in roman and shouldn't contain the original english sentence or any other tag.
English Sentence: <sentence>

---

**10-shot$_\alpha$**

Here are some examples of code-mixed Bangla-English sentences:
<examples>
Convert the following English sentence into code-mixed Bangla-English as in above examples: <English Sentence>

---

## B.2 Dataset Annotation Process

The annotators are experts in both English and an additional language. Being bilingual and multilingual individuals and having participated in code-switching on a daily basis, they were already familiar with the phenomenon of code-mixing. Before the commencement of the annotation process, the phenomenon of code-mixing was explained to them with detailed examples. We had three annotators for English-Hindi and English-Spanish, two annotators for English-French and English Bengali, and one annotator for English-Gujarati.

We gave the following instruction to the annotators: "Please create a code-switched sentence corresponding to the given English sentence"
The annotators were asked to have these criteria met:
1. The code-switched sentences should be grammatically correct.
2. The code-switched sentence must have the same meaning as the corresponding English sentence i.e. it should be an accurate code-mixed translation of the English sentence.

## B.3 Results

### B.3.1 English to Code-Mixed

Figures 3, 4, 5, 6 and 7 graphically depict the BLEU scores achieved for the English to English-Hindi, English-Bengali, English-Gujarati, English-French, and English-Spanish translation tasks respectively.

Figure 8 compares performance across all language pairs.

### B.3.2 Code-Mixed to English

Figures 9, 10, 11, 12 and 13 graphically depict the BLEU scores achieved for the English-Hindi, English-Bengali, English-Gujarati, English-French, and English-Spanish to English translation tasks respectively.

Figure 14 compares performance across all language pairs.

## B.4 Fine-Tuning Setup

### B.4.1 Pre-Processing Data

- Lower Casing was done

- Removed the URLs

- Mentions (starting with @) were removed

- Hashtag symbol (#) was removed but the text was kept

- Characters that appear more than twice in a row were reduced to a single character, including whitespace.

### B.4.2 Fine-Tuning with Twitter Dataset

Total Sentences: 13738 -> Split 0.15 fraction (shuffled) with Valid Data.
Train=11677, Valid=2061
Prefix="Generate Hinglish from English: " was added to input. Max Input Length=114
Max Target Length=118
Optimizer:AdamW
fp16=False
Per Device Train batch size=24
Per Device Eval Batch Size=6
Learning Rate=5e-5
Train Epochs=30
GPU: NVIDIA Tesla P100 (16GB VRAM)

Model with Lowest validation Loss was loaded at the end.

### B.4.3 Inference

Temperature=0.001
Repetition Penalty=2.0

## B.5 Rule Based Prompts' Outputs

> **Output for Rule-1**
>
> 1. Bengali Translation: এটা কি এমন একটি জিজ্ঞাসাবাদী প্রশ্ন?
> 2. PoS Tagging: এটা (PR), কি (PSP), এমন (JJ), একটি (QT), জিজ্ঞাসাবাদী (JJ), প্রশ্ন (NN)?
> 3. List of words: এমন (JJ), জিজ্ঞাসাবাদী (JJ), প্রশ্ন (NN)
> 4. Replaced Bengali sentence: এটা কি such একটি curious question?
> 5. Transliteration: Eta ki such ekti curious question?

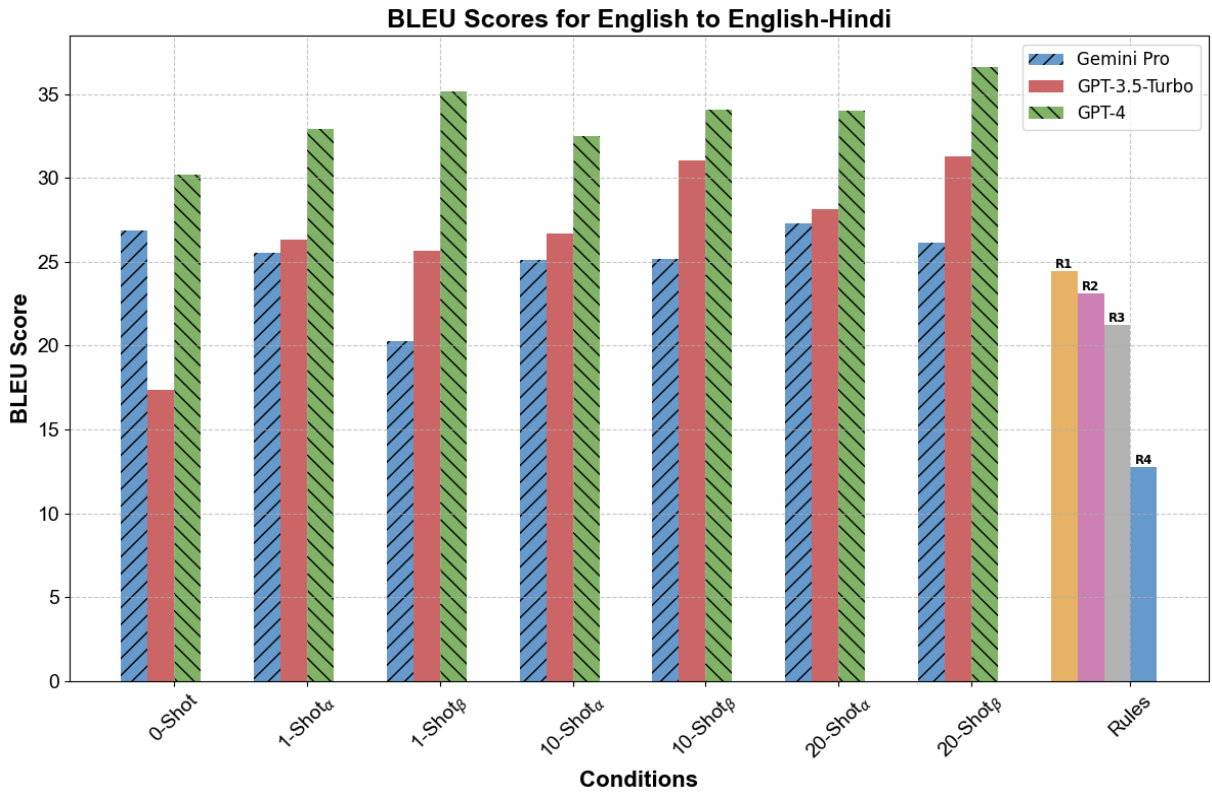

Figure 3: English to English-Hindi

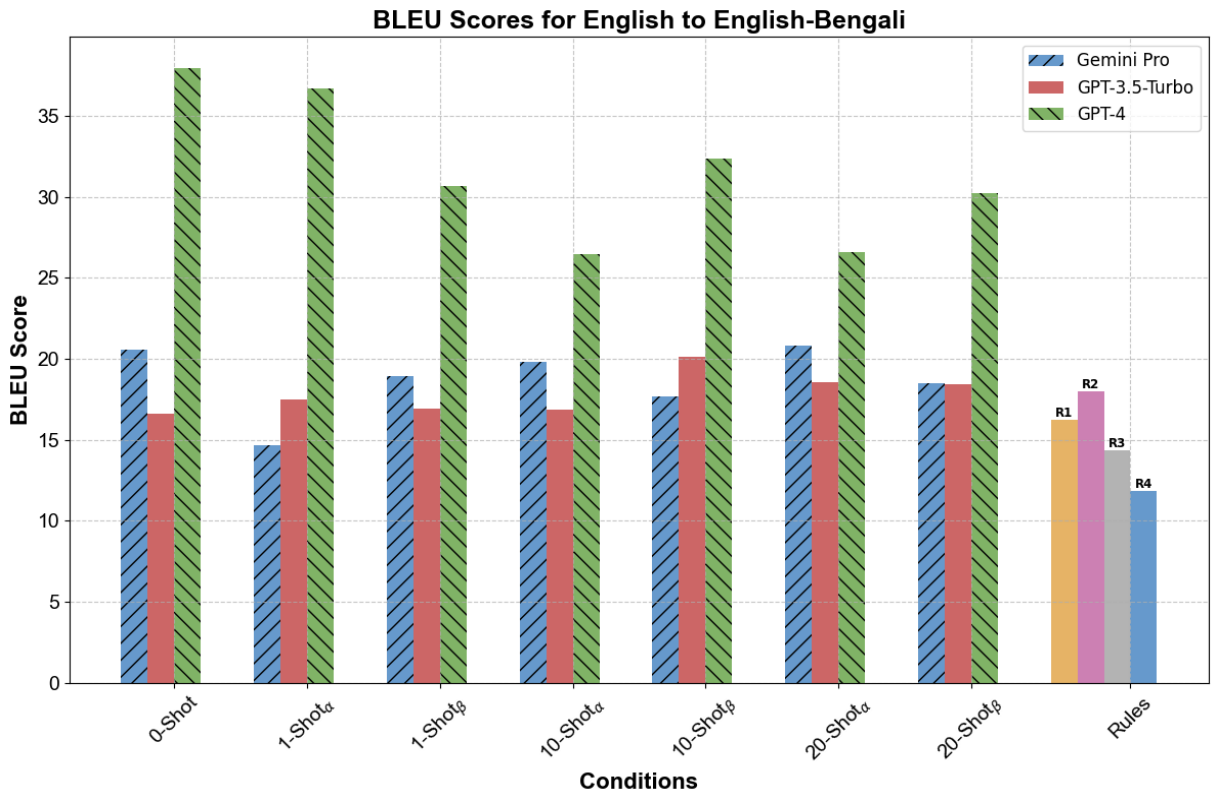

Figure 4: English to English-Bengali

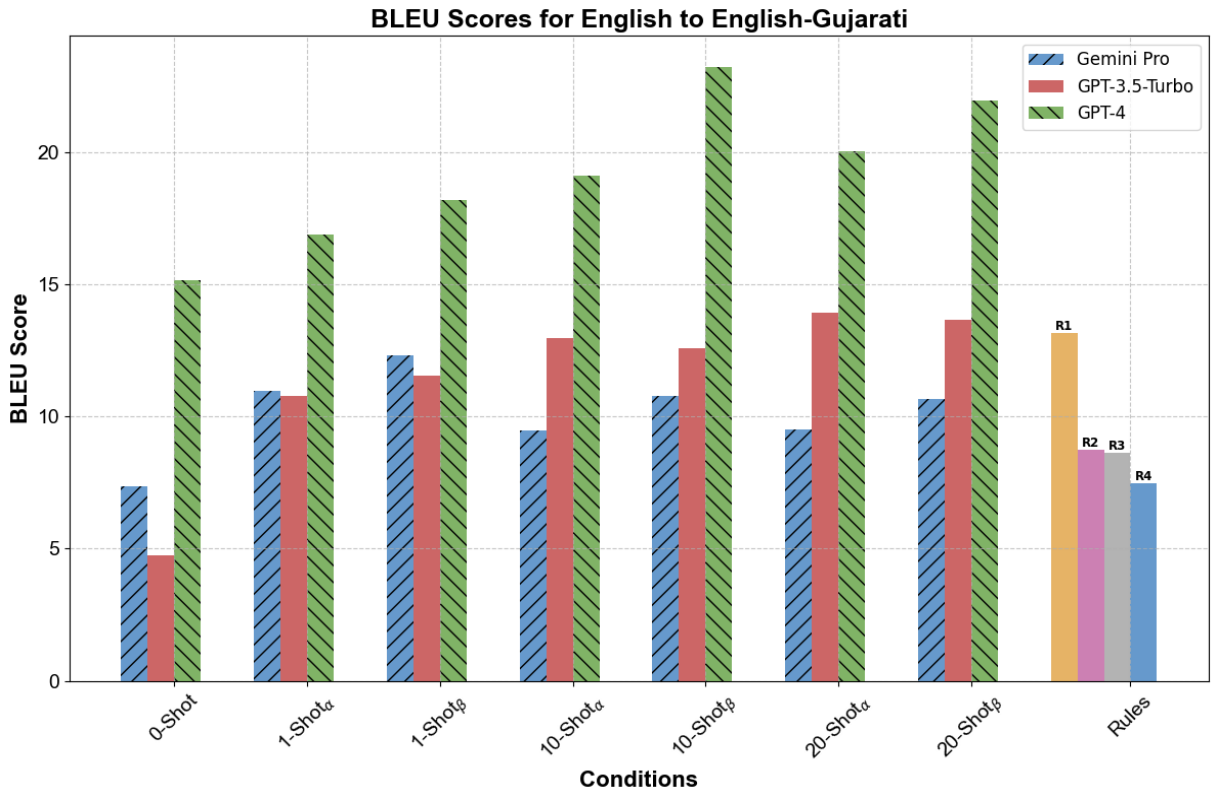

Figure 5: English to English-Gujarati

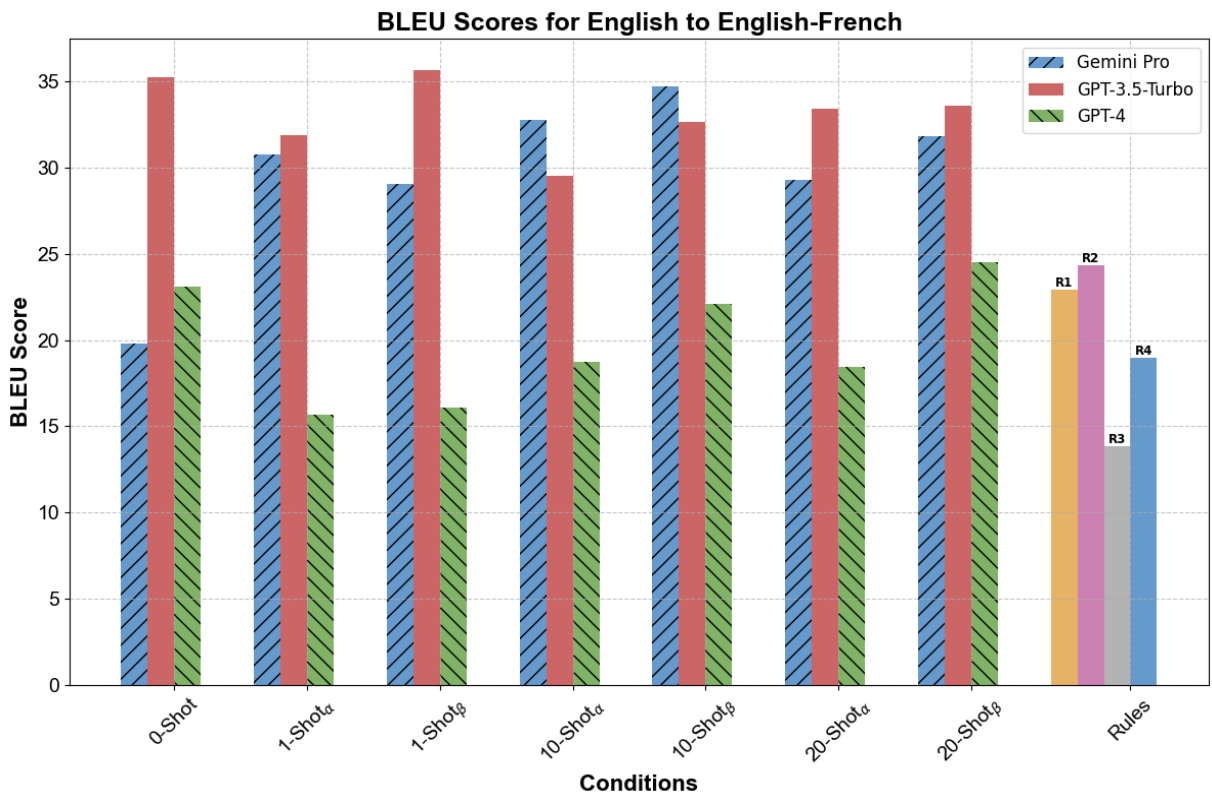

Figure 6: English to English-French

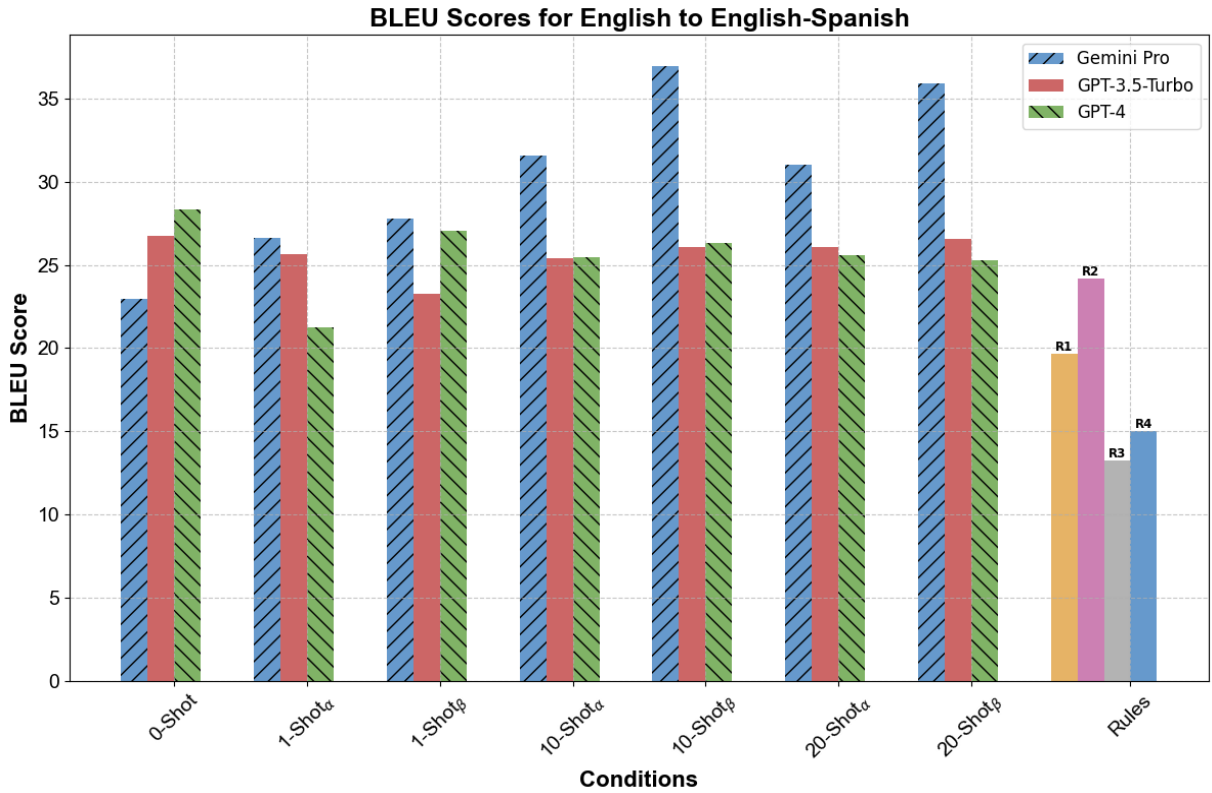

Figure 7: English to English-Spanish

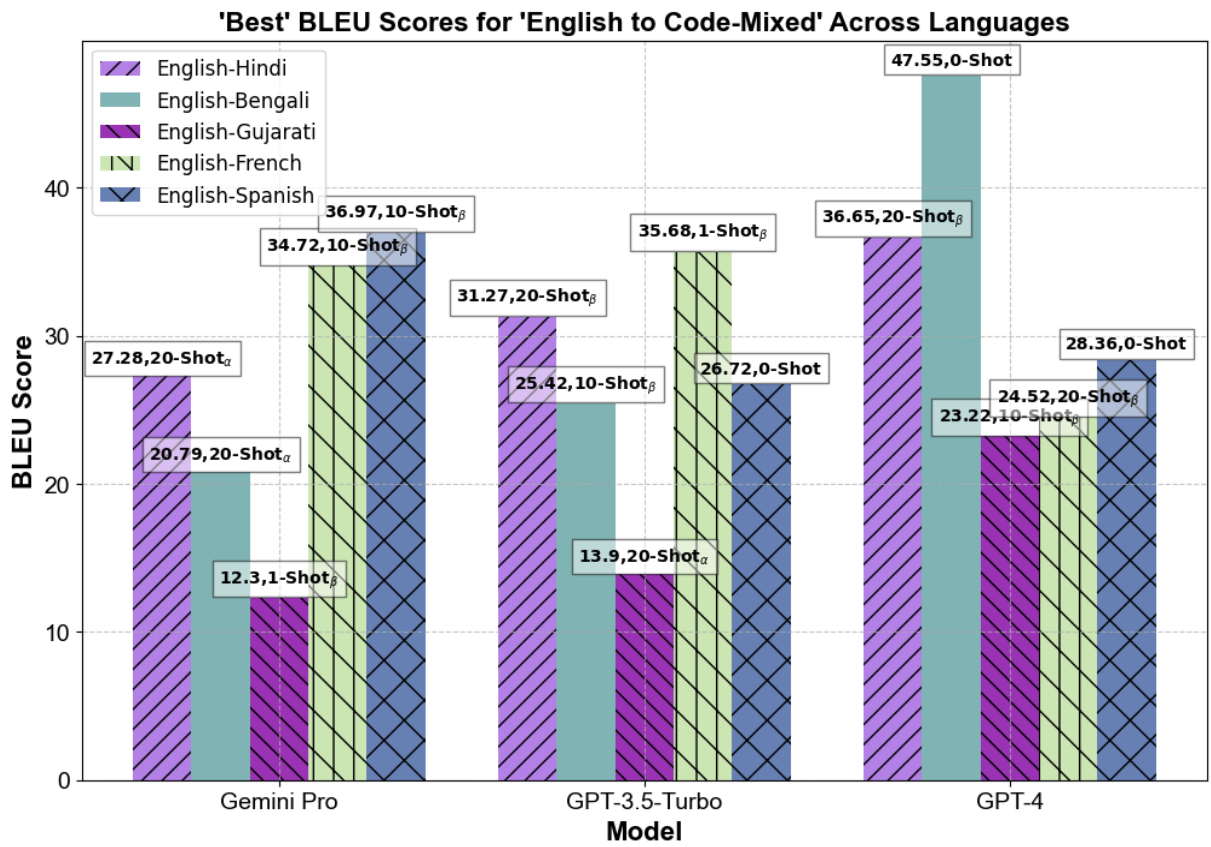

Figure 8: English to Code-Mixed for All Language Pairs

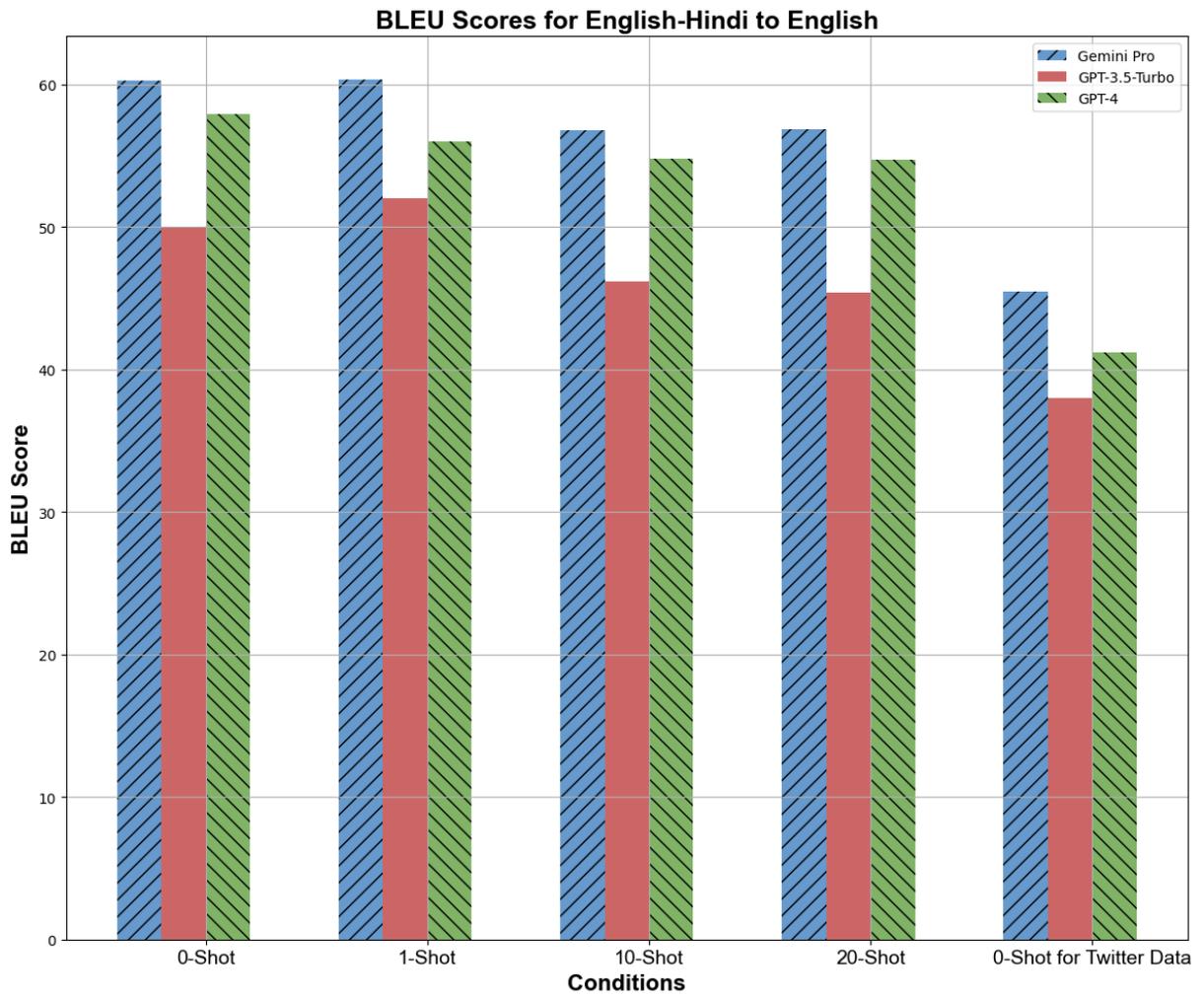

Figure 9: English-Hindi To English

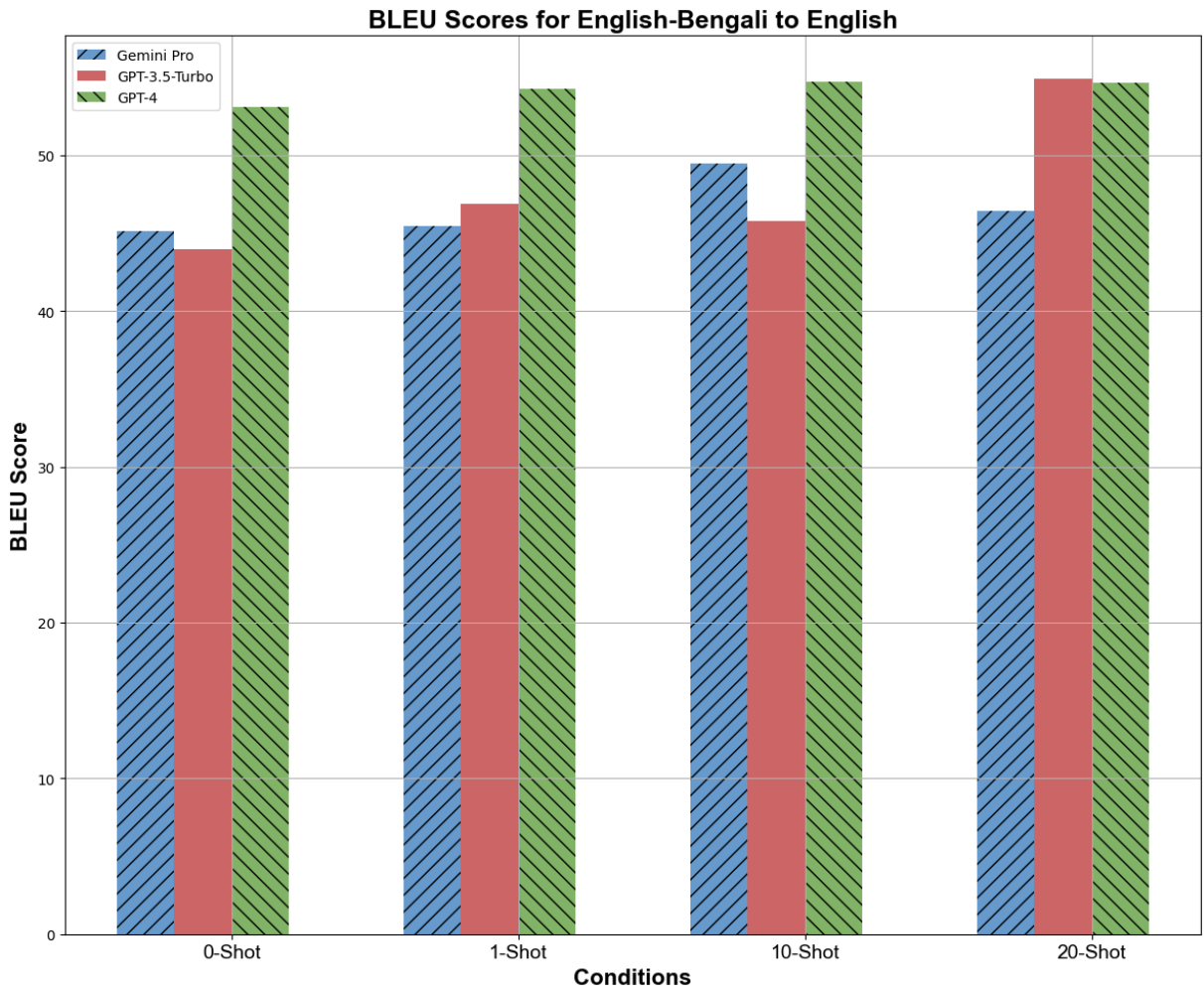

Figure 10: English-Bengali To English

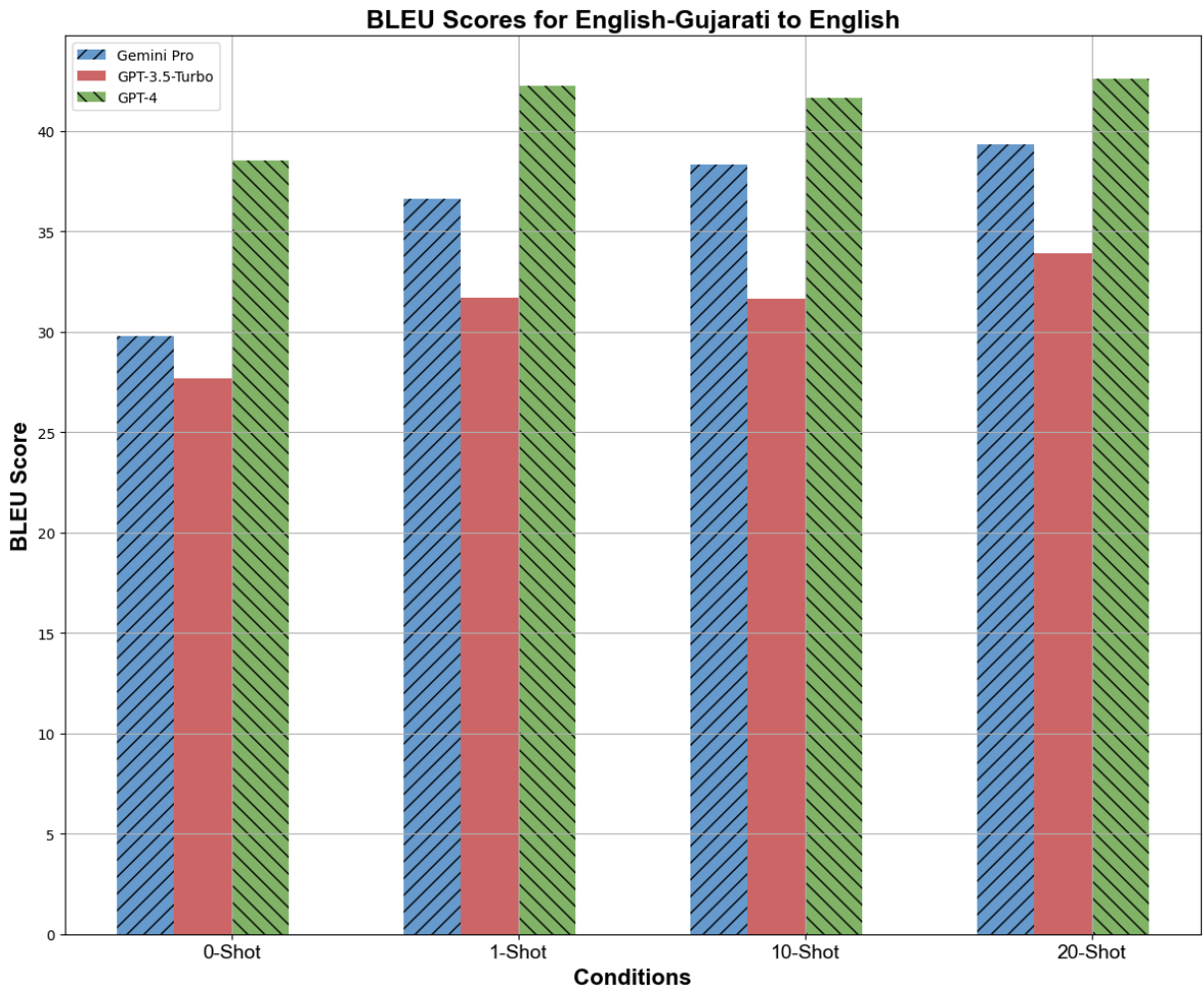

Figure 11: English-Gujarati To English

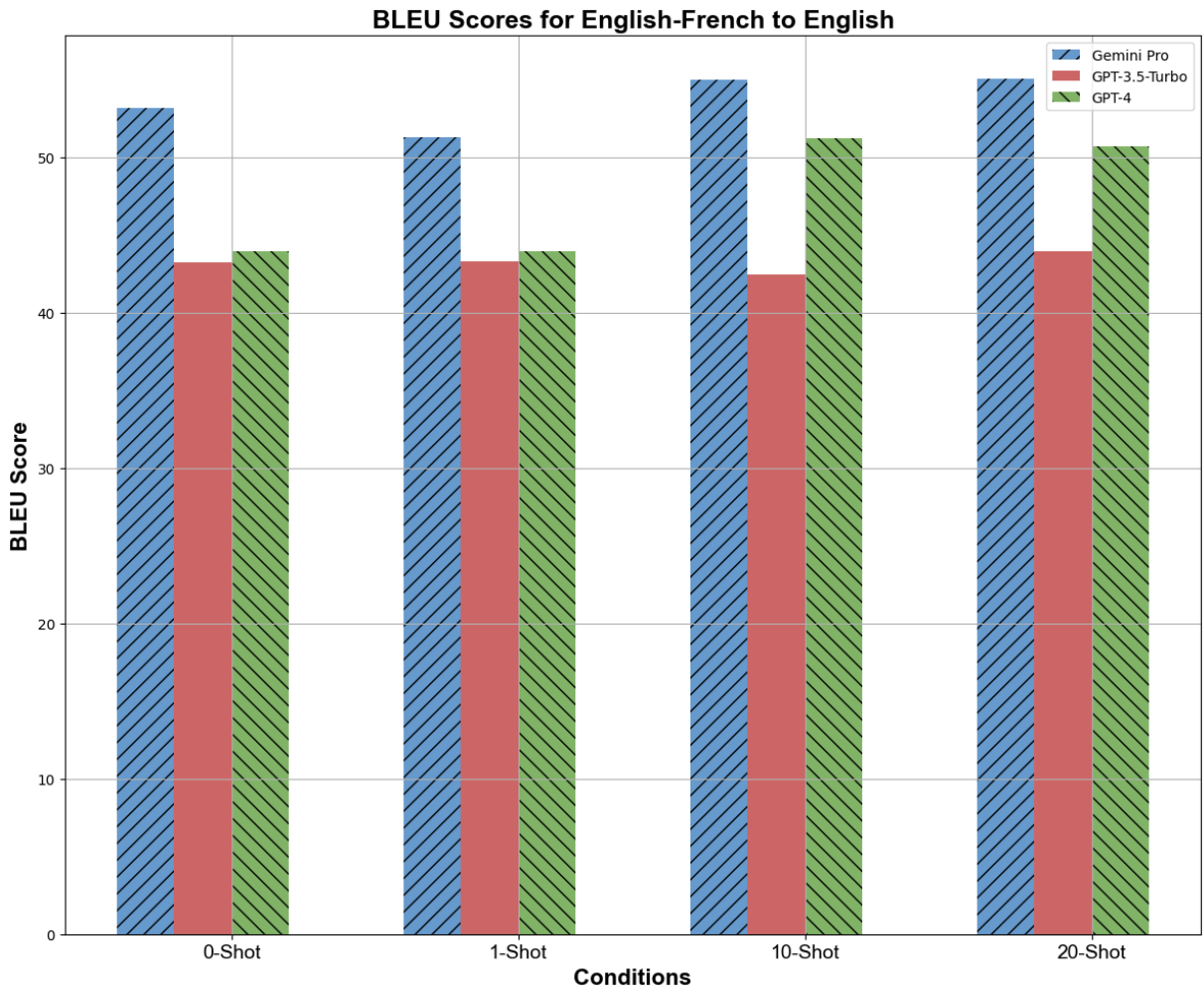

Figure 12: English-French To English

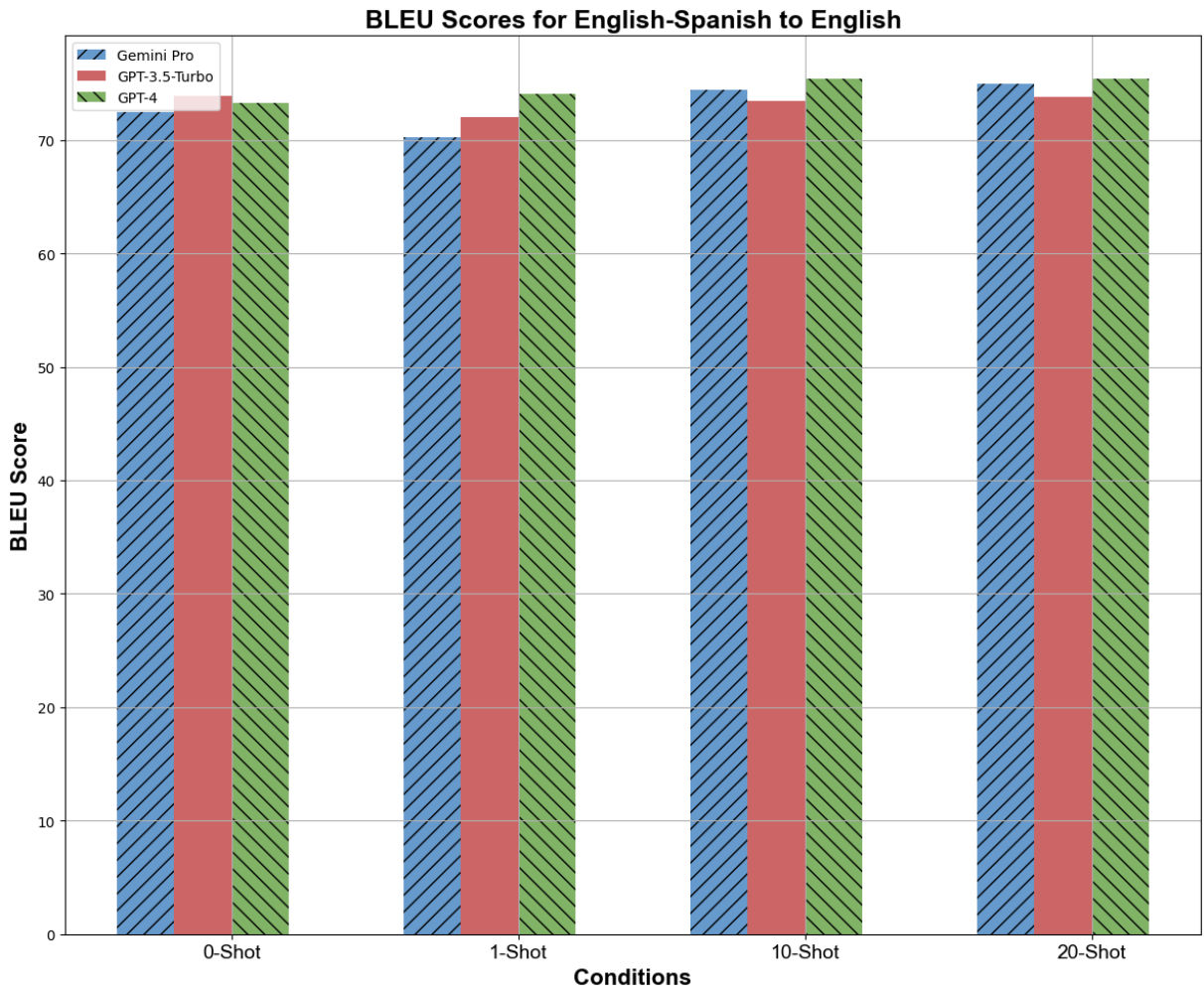

Figure 13: English-Spanish To English

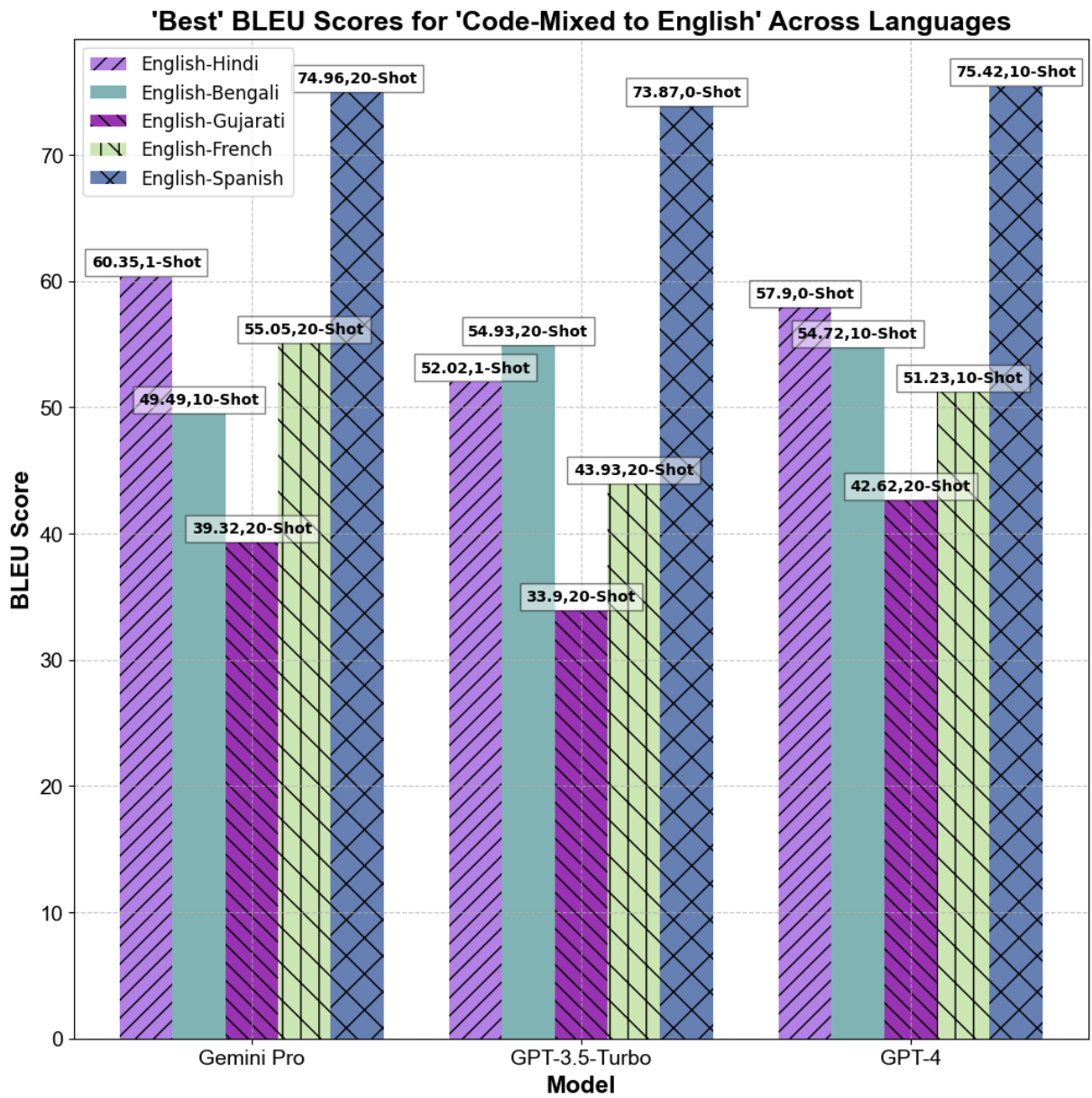

Figure 14: English-Spanish To English

> **Output for Rule-2**
>
> 1. PoS Tagging:
> Is (VBZ) it (PRP) such (JJ) a (DT) curious (JJ) question (NN)?
> 2. List of words which are either Noun (NN), Adjective (JJ), Adverb (RB), CC, or Interjection (UH):
> such, curious, question
> 3. Translation into Bengali:
> এটা কি এমন কৌতূহলী প্রশ্ন?
> 4. Corresponding meaning in Bengali and look them up in the Bengali sentence:
> such - এমন, curious - কৌতূহলী, question - প্রশ্ন
> 5. Replace those words with English counterparts:
> এটা কি such curious question?
> 6. Transliterate the sentence into Roman:
> E ki such curious question?

> **Output for Rule-3**
>
> 1. Noun Phrase Extraction: "a curious question"
> 2. Bengali Translation: এটা কি এমন একটি কৌতূহলপ্রদ প্রশ্ন?
> 3. Corresponding Bengali Phrase: "একটি কৌতূহলপ্রদ প্রশ্ন"
> 4. Replaced Bengali Sentence: এটা কি এমন a curious question?
> 5. Transliteration into Roman: Eta ki emon a curious question?

> **Output for Rule-4**
>
> 1. Phrase Extraction: ["Is it", "such a", "curious question"]
> 2. Second largest length phrase: "such a"
> 3. Bengali Translation: এটা কি এমন একটি কৌতূহলপ্রদ প্রশ্ন?
> 4. Corresponding Bengali words for "such a": "এমন একটি"
> 5. Replacing Bengali phrase with English counterpart: এটা কি such a কৌতূহলপ্রদ প্রশ্ন?
> 6. Transliteration into Roman: Eta ki such a koutuholprod proshno?

### B.6 Output Extraction Prompts

> **Extracting Output from any Step**
>
> You have been given LLM output(in triple ticks) in the format:
> "1. <text>
> 2. <text>
> 3. <text>
> 4. <text>
> 5. <text>: <Sentence>"
>
> Extract the <sentence> in the final output without any tags or anything and make sure that there are no extra inverted commas or '/' sign.
> Just give the final sentence as output It shouldn't be like 'The final sentence is' or anything.
> LLM Output : <Rule Output>

## C Code-Mixed Chatbot

### C.1 Working

The Retrieval component of RAG, which involves Embeddings and the Vector Database, is challenging with code-mixed queries. Our initial experiments with RAG indicate that translating code-mixed query to English is an effective strategy given the current resources (embedding models). Also, as per given a query in English-Bengali, our approach is to first transliterate it in the Bengali script before translating to English. This is in line with out analysis () of Gemini-Pro for English-Bengali to English as well as further experimentation with chatbot. We found that this improves the performance of chatbot for this task.

We first load the document, and perform Hierarchical Chunking into parent nodes of size 2048 and leaf nodes of size 512. The Vector Database is then created and stored. For retrieval, after the query has been translated to English, the embedding is generated for it, and the RAG system generates top twelve relevant or similar chunks. It is to be noted that only Leaf Nodes are considered for this. For a parent node, in case majority of its child nodes are included in these twelve chunks, that parent node is considered instead of these individual child nodes for better context (Auto-Merging). To improve the order of relevance, we use re-ranking and choose the top six chunks. These chunks are passed to the LLM as context along with the query. The answer returned is in English and another prompt is used to generate its code-mixed translation.

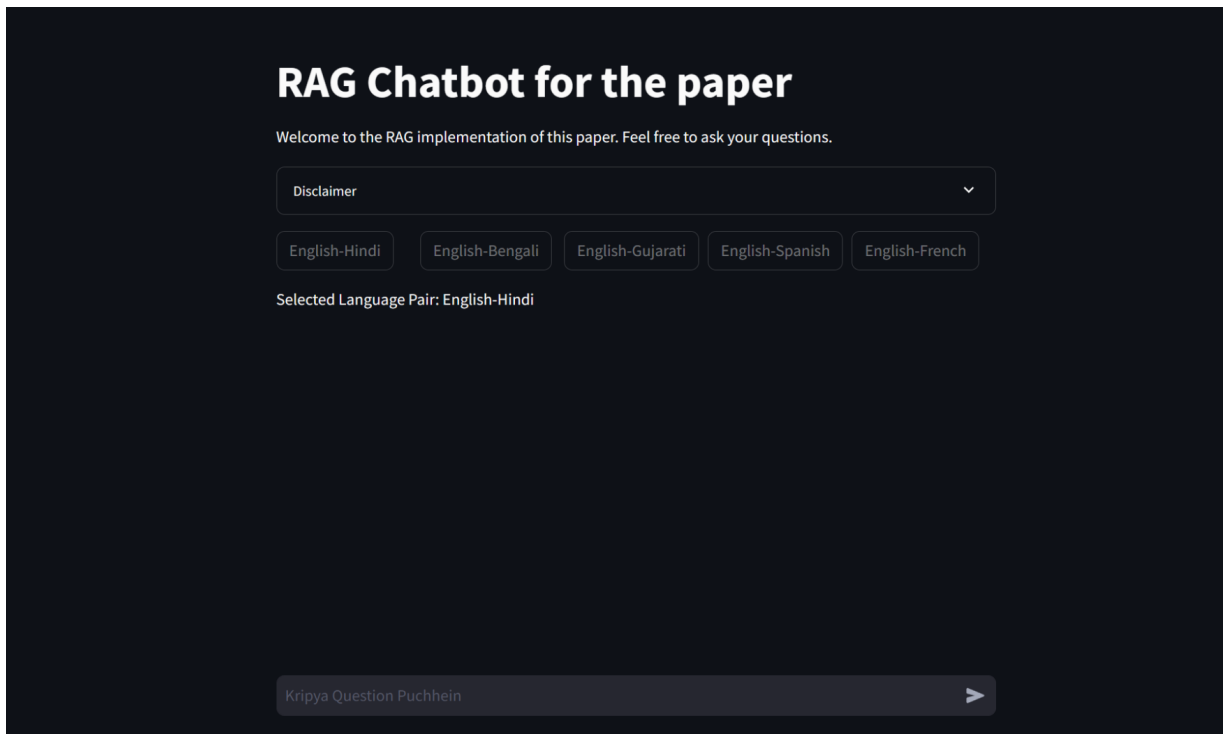

Figure 15: Code-Mixed ChatBot

### C.2 Models and Tools

We use LlamaIndex[4] library. We use Auto-Merging Retrieval Technique from LlamaIndex for RAG. For the embeddings, *BAAI / bge-small-en-v1.5*(Xiao et al., 2023) model has been used.

We use the *BAAI/bge-reranker-base* (Xiao et al., 2023) model for re-ranking.

### C.3 Examples of Chatbot in Operation

We used an incomplete version of this paper to test the Chatbot. Figures 15, 16 and 17 show us some examples.

---

[4]https://www.llamaindex.ai/

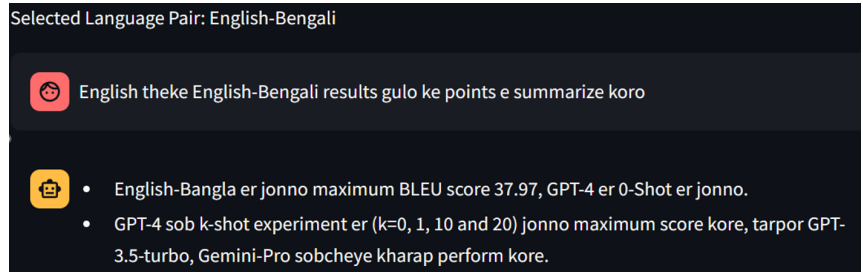

Figure 16: The chatbot has been asked to summarize English to English-Bengali results in points. The question has been asked in English-Bengali. The response talks about the maximum BLEU score that has been achieved for this task and that GPT-4 performs the best in all k-Shot experiments, comparing the performance with the other two models. This seems to be taken from the Results section, where we have also described the results in this way.

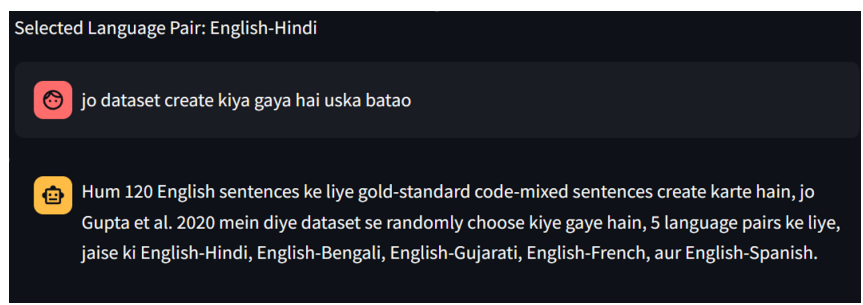

Figure 17: The selected language is English-Hindi.
Translation of User Query : "Tell about the dataset that has been created"
The response is grammatically correct and gives a brief summary about the dataset, mentioning the source of 120 sentences, and the language pairs for which the dataset is.